\pdfoutput=1

\documentclass[11pt]{article}

\usepackage[final]{acl}

\usepackage{times}
\usepackage{latexsym}

\usepackage[T1]{fontenc}

\usepackage[utf8]{inputenc}

\usepackage{microtype}

\usepackage{inconsolata}

\usepackage{graphicx}

\usepackage{amsmath,amssymb,amsfonts}
\usepackage{amsthm}

\usepackage{booktabs}
\usepackage{multicol}
\usepackage{multirow}
\usepackage{color}
\usepackage{xcolor}
\usepackage{subfigure}

\usepackage{algorithm}
\usepackage{algorithmic}

\usepackage{array} 
\usepackage{makecell}

\title{\textsc{ChatDify}: Automating End-to-End Dify Workflow Synthesis from Natural Language via Multi-Agent Collaboration}

\author{
 \textbf{Tao Fan\textsuperscript{1, 2}},
 \textbf{Guoqiang Ma\textsuperscript{2}}
\\
 \textsuperscript{1} Hong Kong University of Science and Technology, Hong Kong, China
 \\
 \textsuperscript{2}WeBank Co., Ltd, Shenzhen, China
\\
 \small{
   \textbf{Correspondence:} \href{mailto:tfanac@cse.ust.hk}{tfanac@cse.ust.hk}
 }
\\
 \small{
   \textbf{Code:} \url{https://github.com/OpenAgentis/chatdify/}
 }
}

\begin{document}
\maketitle

\begin{abstract}
The proliferation of Large Language Models has catalyzed powerful application development platforms like Dify, which enable users to build sophisticated LLM-powered workflows through modular node orchestration. However, manually configuring these workflows requires deep platform expertise and remains time-consuming and error-prone, creating a significant barrier for non-expert users. To address this challenge, we propose \textsc{ChatDify}, a novel multi-agent framework that automates end-to-end Dify workflow synthesis from natural language descriptions. \textsc{ChatDify} employs an Actor-Critic architecture comprising three specialized agents: the \textit{Planner Agent} that decomposes user requests into structured execution plans, the \textit{Executor Agent} that performs context-aware parameter synthesis and schema-augmented DSL compilation, and the \textit{Reviewer Agent} that ensures reliability through hybrid verification with closed-loop refinement. We evaluate \textsc{ChatDify} on a curated benchmark of 100 production-inspired tasks across three state-of-the-art LLM backends, achieving  0.96, 0.91, and 0.75 success rates with Qwen-3.7, GLM-5.1, and DeepSeek-v4, respectively. These results demonstrate its robustness and practical applicability for lowering the technical barrier to creating sophisticated LLM applications. 
\end{abstract}

\section{Introduction}
The proliferation of Large Language Models (LLMs)  ~\cite{Chatgpt, OpenAI2023GPT4TR, comanici2025gemini, dubey2024llama, deepseekai2026deepseekv4,zeng2026glm,yang2025qwen3} has catalyzed the emergence of powerful application development platforms like Dify~\cite{dify_ai_2023}, Coze~\cite{coze_ai_2024} and LangChain~\cite{mavroudis2024langchain}. These platforms provide developers with modular tools to build sophisticated 
\begin{figure}[h]
  \centering
 \includegraphics[width=1.0\linewidth]{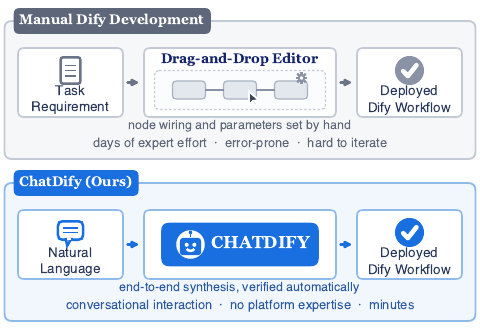} 
  \caption{\textsc{ChatDify} Automates Dify Workflow Synthesis via Natural Language, Advancing Beyond Manually Designed Drag-and-Drop Dify Workflows.}
  \label{fig:chatdify-vs-traditional}
\end{figure}
LLM-powered applications, such as Retrieval-Augmented Generation (RAG) systems~\cite{lewis2020retrieval,gao2023retrieval}, chatbots~\cite{adamopoulou2020overview}, and 
complex agentic workflows. Dify, in particular, provides a user-friendly interface to facilitate the orchestration of various nodes including LLMs, knowledge bases, and custom tools into a coherent workflow structure, as illustrated in Figure~\ref{fig:dify_deepresearch_demo}.

Despite the power of these platforms, a significant barrier to entry remains. Users must manually select, connect, and configure each node in the workflow, a process that demands both a clear understanding of the desired logic and familiarity with the platform's specific functionalities. This manual configuration is time-consuming, error-prone, and can be particularly daunting for users without a technical background. As workflows grow in complexity, the cognitive load on developers increases, hindering rapid prototyping and iteration.

A natural remedy is to let users describe their goals in natural language and generate the workflow automatically. However, this is far from trivial, because a generated workflow must be correct at three levels, and each level can fail even when the previous one holds.
(i) \textbf{Structural correctness}: the request must be decomposed into a valid DAG, with appropriate nodes selected from Dify's heterogeneous node catalog and arranged with correct branching and iteration logic.
(ii) \textbf{Configuration correctness}: each node must follow Dify's platform-specific schema, and its variable references (e.g., \texttt{\{\{node\_id.field\}\}}) must point to type-compatible upstream outputs; general-purpose LLMs do not reliably master these details, and an error in one node can break all downstream nodes.
(iii) \textbf{Semantic correctness}: a valid and executable workflow may still diverge from the user's intent, and its open-ended outputs lack deterministic ground truth to check against.

To address these challenges, we propose \textsc{ChatDify}, a novel framework that enables users to synthesize and configure complete Dify workflows using only natural language. By simply describing their goal, users can leverage \textsc{ChatDify} to automatically construct the corresponding workflow graph, populate node parameters, and establish the necessary connections. This transforms the development process from a manual, designed drag-and-drop Dify workflow into a simple conversational experience, as illustrated in Figure~\ref{fig:chatdify-vs-traditional}.

\textsc{ChatDify} orchestrates a multi-agent collaborative framework to synthesize executable Dify workflows from natural language descriptions (see Figure~\ref{fig:chatdify_architecture}). This architecture comprises three specialized agents aligned with these three levels: the \textbf{Planner Agent} targets structural correctness by deconstructing high-level user requests into a structured, step-by-step execution plan; the \textbf{Executor Agent} targets configuration correctness through context-aware parameter synthesis and schema-augmented DSL compilation using a dedicated node knowledge base; and the \textbf{Reviewer Agent} verifies all three levels by combining static symbolic verification with dynamic runtime probing, driving a closed-loop refinement process. This integrated process substantially improves the reliability of generated workflows.
The main contributions of this work are summarized as follows:

\begin{itemize}
   \item \textbf{End-to-End NL-to-Dify Synthesis System.} We present \textsc{ChatDify}, a framework for translating natural-language requirements into executable Dify workflows through workflow graph construction, node-level parameter synthesis, and schema-aware DSL compilation.

    \item \textbf{Actor-Critic Architecture with Hybrid Verification.} We design a knowledge-grounded Planner--Executor--Reviewer pipeline where the Reviewer combines static checks over schema, topology, and variables with dynamic probing that pairs sandbox execution and an \emph{LLM-as-Judge} semantic verifier.

    \item \textbf{Industrial Benchmark and Empirical Evidence.} We curate \textbf{100 production-inspired Dify tasks} across 8 categories. Across Qwen-3.7, GLM-5.1, and DeepSeek-v4, \textsc{ChatDify} achieves \textbf{0.96}, \textbf{0.91}, and \textbf{0.75} success rates, substantially outperforming ZeroShot and ZeroShot+Dynamic baselines.
\end{itemize}

\section{Related Work}

\subsection{NL2Workflow}
The task of generating executable workflows from natural language has advanced rapidly with LLMs. AutoFlow~\cite{li2024autoflow} combines in-context learning with supervised fine-tuning and reinforcement learning to synthesize workflow plans, WorkflowLLM~\cite{fan2024workflowllm} fine-tunes an LLM on large workflow corpora, and Action Engine~\cite{esashi2025action} generates Function-as-a-Service (FaaS) pipelines with tool-augmented LLMs. In industrial settings, WorkTeam~\cite{liu2025workteam} divides workflow construction among supervisor, orchestrator, and filler agents, while Classifier-Augmented Generation (CAG)~\cite{gschwind2025classifier} builds IBM DataStage ETL workflows through classifier-guided stage prediction, edge prediction, and property inference. These methods either operate at a higher level of abstraction or target other platforms, and do not address Dify-specific node configuration. \textsc{ChatDify} instead synthesizes executable Dify workflows with node-level parameters, explicit data-flow connections, and cross-node variable references, and verifies each candidate through static checks and sandboxed execution.

\subsection{LLM Agents}

LLM-powered agents have emerged as a compelling paradigm for systems that can reason, plan, and act toward complex goals. Prior work spans three complementary lines. (i) \textit{Tool-augmented single-agent} methods enable an LLM to call external tools and interleave reasoning and acting, exemplified by ReAct~\cite{yao2022react} and Toolformer~\cite{schick2023toolformer}. (ii) \textit{Self-reflection and iterative improvement} strategies, such as Reflexion~\cite{shinn2023reflexion}, let agents monitor outcomes and refine plans or prompts to improve reliability. (iii) \textit{Multi-agent collaboration} frameworks organize specialized roles (planners, executors, and critics) into coordinated protocols for complex tasks; representative systems include AutoGen~\cite{wu2024autogen}, CodeAct~\cite{wang2024executable}, CAMEL~\cite{li2023camel}, MetaGPT~\cite{hong2023metagpt}, and Voyager~\cite{wang2023voyager}. \citet{ijcai2024p890} survey LLM-based multi-agent systems, covering their application settings, agent profiling and communication, and capability acquisition. \textsc{ChatDify} follows the multi-agent line with specialized planner, executor, and critic roles that communicate through structured plans and diagnostic feedback.

\begin{figure*}[h]
  \centering
 \includegraphics[width=1.0\linewidth]{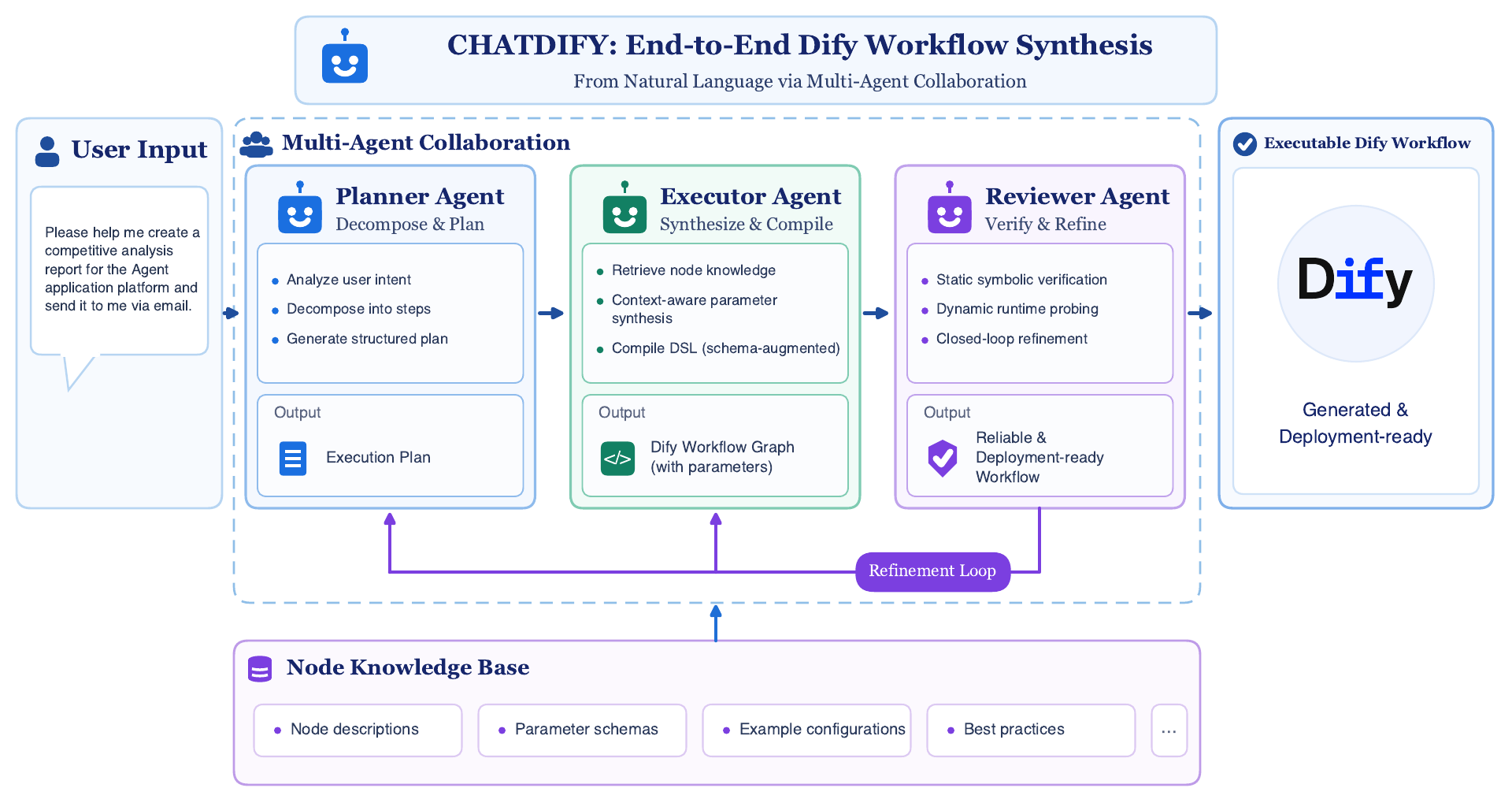} 
\caption{Overview of \textsc{ChatDify} Architecture: A multi-agent system for Dify workflow synthesis. The system employs an Actor-Critic paradigm where the Planner and Executor (Actors) synthesize the workflow, while the Reviewer (Critic) provides hybrid feedback for closed-loop refinement.}
\label{fig:chatdify_architecture}
\end{figure*}

\section{The Proposed \textsc{ChatDify} Framework}
\label{sec:framework}

\textsc{ChatDify} automates the synthesis of executable Dify workflows from natural language requirements. Formally, we define a Dify workflow $W$ as a triple $W = (N, E, \Theta)$, where $N = \{n_1, \dots, n_k\}$ denotes a set of heterogeneous functional nodes (e.g., LLM, Code), $E \subseteq N \times N$ represents directed data-flow dependencies forming a Directed Acyclic Graph (DAG), and $\Theta = \{\theta_1, \dots, \theta_k\}$ specifies the dify-specific parameter configurations for each node. 

As illustrated in Figure~\ref{fig:chatdify_architecture}, \textsc{ChatDify} adopts an \textbf{Actor-Critic architecture}~\cite{konda1999actor,lowe2017multi} comprising three specialized agents: \textit{Planner}, \textit{Executor}, and \textit{Reviewer}. The synthesis is modeled as a closed-loop optimization where agents collaboratively: (i) map user requests to a topological plan, (ii) perform context-aware synthesis and schema-augmented compilation into Dify's Domain-Specific Language (DSL), and (iii) execute hybrid verification to drive closed-loop refinement.

\subsection{Planner Agent: Hierarchical Decomposition and Pattern-Augmented Planning}

Given a user's request $U$, the \textbf{Planner Agent} serves as the high-level architect, aiming to produce a structured execution plan $P = (T, E)$, where $T = \{t_1, t_2, \dots, t_k\}$ is a set of sub-tasks and $E$ is a set of directed edges representing data dependencies between these tasks. This process is formulated as:
\begin{equation}
    P = \mathcal{F}_{\text{Planner}}(U, K)
\end{equation}
where $K$ is the Dify Knowledge Base. 

To ensure scalability and logical consistency, the Planner employs a \textbf{top-down hierarchical decomposition and planning} strategy:

\begin{itemize}
    \item \textbf{Intent Recognition and Block Mapping}: Upon receiving $U$ (e.g., ``generate a competitive analysis report and email it''), the Planner parses the semantic intent to perform a \textbf{functional decomposition}, breaking the high-level goal into coarse-grained functional blocks (e.g., `\texttt{Data Collection}' $\rightarrow$ `\texttt{Information Processing \& Analysis}' $\rightarrow$ `\texttt{Report Generation}' $\rightarrow$ `\texttt{Result Delivery}').

    \item \textbf{Pattern-Augmented Reference Retrieval}: To avoid designing from scratch, the Planner performs semantic retrieval over $K$ to find the most relevant \textbf{workflow sub-graph templates}. This pattern-augmented approach ensures the generated plan inherits industry best practices (e.g., RAG pipelines or multi-hop reasoning chains).

    \item \textbf{DAG Synthesis and Node Selection}: In the final planning stage, the Planner instantiates the abstract functional blocks into the concrete sequence of executable sub-tasks $T$ by leveraging retrieved templates and specific request constraints. For each sub-task, the Planner selects the optimal Dify functional nodes from $K$ (e.g., ``Tavily Search'', ``Knowledge Retrieval'', ``LLM''). Simultaneously, it explicitly defines the data-flow dependencies $E$ to connect these nodes, thereby synthesizing the final topological DAG plan $P$.
    
\end{itemize}
This plan $P$ acts as a precise ``\textit{blueprint},'' providing unambiguous logical specifications for the subsequent context-aware synthesis phase. As part of the closed-loop optimization process, the Planner is also responsible for re-planning and adjusting the plan $P$ based on high-level logical feedback from the Reviewer Agent.

\subsection{Executor Agent: Context-Aware Synthesis and Schema-Augmented Compilation}

The \textbf{Executor Agent} acts as the system's ``compiler,'' tasked with synthesizing node parameters and compiling the abstract logical plan $P$ into a platform-specific, concrete, and executable Dify workflow $W$. 
This process is defined as:
\begin{equation}
    W = \mathcal{F}_{\text{Executor}}(P, U, K)
\end{equation}
where $W$ consists of instantiated nodes $N = \{n_1, \dots, n_k\}$, their specific parameter configurations $\Theta = \{\theta_1, \dots, \theta_k\}$, and the connections between them ($E$). $K$ provides the schema specifications for Dify nodes. 

The key to the Executor's operation lies in \textbf{context-aware synthesis} and \textbf{schema-augmented compilation}, which involves two core steps:

\begin{itemize}
  \item \textbf{Context-Aware Parameter Synthesis}: 
    For each sub-task $t_i \in T$, the Executor instantiates the corresponding Dify node type from $K$. The agent performs \textbf{generative parameter synthesis} to populate configuration $\theta_i$. This synthesis is conditioned not only on the user request $U$ and plan $P$, but critically on the \textbf{upstream execution context} (e.g., outputs from predecessor nodes $\{n_1, \dots, n_{i-1}\}$). This allows for the dynamic construction of complex attributes, such as synthesizing task-specific system prompts for ``LLM'' nodes or executable logic for ``Code'' nodes. For underspecified parameters, a context-aware defaulting mechanism completes the configuration based on node functional semantics.

    \item \textbf{Schema-Augmented DSL Compilation}: The Executor finally assembles all instantiated nodes and their edge connections into a unified YAML structure following the Dify DSL syntax. To enforce syntactic validity and reduce configuration errors, it applies a \textbf{schema-augmented synthesis} strategy. This mechanism restricts the generation process to strictly adhere to the \textbf{multi-dimensional protocol specifications} defined in $K$, spanning key constraints such as: (i) \textit{Node-Specific Schemas}, which dictate the mandatory and optional field structures for each node type, alongside their associated data types and range boundaries; (ii) \textit{Variable Propagation Protocols}, validating the syntactic correctness and path-traceability of cross-node references (e.g., \texttt{\{\{node\_id.field\}\}}).
    
\end{itemize}
The Executor Agent completes the critical synthesis and compilation from a logical task plan into an executable workflow instance on the Dify platform, effectively turning the ``\textit{design blueprint}'' into ``\textit{runnable code}.'' Within the closed-loop optimization process, the Executor receives specific error feedback from the Reviewer Agent regarding node parameters, configurations, or connections, and performs targeted repairs and re-instantiation.

\subsection{Reviewer Agent: Hybrid Verification and Closed-loop Refinement}

The \textbf{Reviewer Agent} serves as the Critic, validating $W$ through a hybrid process of static symbolic analysis and dynamic runtime verification:
\begin{equation}
    F = \mathcal{F}_{\text{Reviewer}}(W) = \mathcal{F}_{\text{dynamic}}(\mathcal{F}_{\text{static}}(W))
\end{equation}
where $F$ is a set of feedback signals. If $F \neq \emptyset$, the framework enters an \textbf{Iterative Repair Loop}: $W^{(t+1)} = \text{Refine}(W^{(t)}, F)$, continuing until the workflow satisfies all constraints.

\subsubsection{Static Symbolic Verification}
The Reviewer performs a rigorous symbolic analysis without executing the workflow:
\begin{itemize}
    \item \textbf{Topological Integrity}: Ensures the graph is a valid DAG by detecting cycles, unreachable nodes, or broken data-flow edges, thereby guaranteeing a feasible execution order.
    \item \textbf{Type-Safe Variable Propagation}: Constructs a \textbf{variable dependency graph} to trace all cross-node references (e.g., \texttt{\{\{node\_id.field\}\}}). It verifies that every referenced variable is properly defined in an upstream node and that its data type matches the expected input schema of the consuming node.
    \item \textbf{Schema Compliance}: Validates the YAML against Dify's official schema, checking for missing required fields, parameter type mismatches, and out-of-range values.
\end{itemize}

\subsubsection{Dynamic Runtime Probing}
Static analysis cannot expose runtime-only defects, nor judge whether a successfully executed workflow actually fulfills the user's intent. Once $W$ passes static checks, the Reviewer runs a two-stage probe in a \textbf{controlled Dify sandbox}:
\begin{itemize}
    \item \textbf{Sandboxed Execution Probing}: Auto-synthesizes representative test inputs from $U$ and $W$, executes $W$ end-to-end, and collects per-node status, traces, and outputs to surface \textit{hard execution failures} (e.g., node exceptions, unresolved references, tool/API errors, missing credentials, timeouts).
    \item \textbf{LLM-as-Judge Semantic Verification}: Since Dify workflows are open-ended (generation, summarization, tool use) and lack deterministic ground truth, an \textbf{LLM judge} consumes $(U, W, \text{trace}, \text{output})$ and returns a structured verdict with a node-level rationale, flagging cases where execution succeeds but behavior diverges from $U$.
\end{itemize}

\section{Experiments}
\label{sec:experiments}

\subsection{Experimental Setup}
\label{ssec:setup}

\textbf{Benchmark Dataset.} We construct 100 production-inspired Dify workflow tasks covering eight categories, as shown in Table~\ref{tab:task_distribution}. These tasks are abstracted from common Dify application scenarios and workflow patterns.

\begin{table}[h]
    \centering
    \footnotesize
    \setlength{\tabcolsep}{3pt}
    \begin{tabular}{lcc}
        \toprule
        \textbf{Task Category} &  \textbf{Count} & \textbf{Percentage} \\
        \midrule
        Text Generation Tasks & 18& 18\%\\
        Information Extraction & 18& 18\%\\
        Multimodal Understanding & 11 & 11\% \\
        Multimodal Generation & 8& 8\%\\
        Data Analysis and Processing & 14& 14\%\\
        Conversation and Interaction & 10& 10\%\\
        Code and Programming & 10& 10\%\\
        Research and Application & 11& 11\%\\
        \midrule
        \textbf{Total} & \textbf{100} & \textbf{100\%} \\
        \bottomrule
    \end{tabular}
    \caption{Task distribution across eight categories in the benchmark.}
    \label{tab:task_distribution}
\end{table}

\textbf{Models and Baselines.} 
We instantiate \textsc{ChatDify} with \textbf{Qwen-3.7 max}~\cite{yang2025qwen3}, \textbf{GLM-5.1}~\cite{zeng2026glm}, and \textbf{DeepSeek-v4 pro}~\cite{deepseekai2026deepseekv4}.  We compare with \textbf{ZeroShot} and \textbf{ZeroShot+Dynamic}, and ablate static verification, dynamic runtime probing, and their combination. For fair comparison, both \textsc{ChatDify} and ZeroShot+Dynamic run for up to three repair epochs by default.

\textbf{Evaluation Protocol and Metrics.} 
We report \textbf{Success Rate (SR)} and \textbf{Token}, the average token consumption in thousands. A task counts as successful if the workflow returned within the epoch budget passes three sequential checks: (i) a \textit{static check}, requiring schema compliance, a valid DAG topology, and correctly defined variable references; (ii) a \textit{runtime check}, where the workflow is imported into the Dify sandbox and executed end-to-end on a test case without node exceptions, unresolved references, tool/API errors, or timeouts; and (iii) a \textit{semantic check}, where an LLM judge receives the request, the workflow DSL, the test case, the final outputs, and a per-node execution summary, and returns a binary verdict. The judge is the semantic verifier of the Reviewer Agent and uses the same backbone LLM as the generator. All methods, including ZeroShot and ZeroShot+Dynamic, are scored with this protocol.

\textbf{Implementation Details.} 
All experiments use Dify \texttt{v1.11.2} and LLM APIs from Volcengine (GLM-5.1), Deepseek-AI (DeepSeek-v4 pro) and Alibaba Cloud (Qwen-3.7 max), running on a 4-core Intel(R) Xeon(R) Platinum CPU server with 16 GB RAM.

\begin{table}[ht]
    \centering
    \footnotesize 
    \setlength{\tabcolsep}{2.5pt} 
    \begin{tabular}{llcc}
        \toprule
        \textbf{Model} &\textbf{Method} &\textbf{ SR  }& \textbf{Token (k)} \\

         \midrule
        \multirow{3}{*}{Qwen-3.7} 

         & ZeroShot        & 0.43& 24\\
        
        & ZeroShot+Dynamic        & 0.67& 52\\
        
        & \textbf{\textsc{ChatDify} (ours)}  & \textbf{0.96}& 256\\
        
        \midrule
        \multirow{3}{*}{GLM-5.1} 
         & ZeroShot        & 0.13& 13\\
        
        & ZeroShot+Dynamic        & 0.45& 36\\
        
        & \textbf{\textsc{ChatDify} (ours)} & \textbf{0.91}& 348\\

        \midrule
        \multirow{3}{*}{DeepSeek-v4} 

        & ZeroShot        & 0.15& 15\\
        
        & ZeroShot+Dynamic        & 0.36& 40\\
        
        & \textbf{\textsc{ChatDify} (ours)}   & \textbf{0.75}& 385\\
        
        \bottomrule
    \end{tabular}
    
     \caption{Main comparison across LLM backends and practical baselines.}
     \label{tab:model_evaluation}
\end{table}

\begin{table}[ht]
    \centering
    \footnotesize 
    \setlength{\tabcolsep}{2.5pt} 
    \begin{tabular}{llcc }
        \toprule
        \textbf{Model} &\textbf{Method} &\textbf{ SR }& \textbf{Token (k)} \\

         \midrule
        \multirow{4}{*}{Qwen-3.7}& \textbf{Full \textsc{ChatDify}}  & \textbf{0.96}& 256\\
        & w/o Static         & 0.93& 270\\
        & w/o Dynamic        & 0.73& 175\\
        & w/o Static \& Dynamic        & 0.71& 166\\

        \midrule
        \multirow{4}{*}{GLM-5.1}& \textbf{Full \textsc{ChatDify}}  & \textbf{0.91}& 348\\
        & w/o Static         &0.83& 426\\
        & w/o Dynamic       & 0.71& 219\\
        & w/o Static \& Dynamic        & 0.60& 210\\

        \midrule
        \multirow{4}{*}{DeepSeek-v4}& \textbf{Full \textsc{ChatDify}}  & \textbf{0.75}& 385\\
        & w/o Static         & 0.67& 397\\
        & w/o Dynamic        & 0.44& 162\\
        & w/o Static \& Dynamic        & 0.33& 145\\

        \bottomrule
    \end{tabular}
    
     \caption{Ablation results on static verification and dynamic runtime probing.}
     \label{tab:model_ab_sd}
\end{table}

\begin{table}[ht]
    \centering
    \footnotesize 
    \setlength{\tabcolsep}{2.5pt} 
    \begin{tabular}{llcc }
        \toprule
        \textbf{Model} &\textbf{Epochs} &\textbf{ SR  }& \textbf{Token (k)} \\
     
         \midrule
        \multirow{3}{*}{Qwen-3.7}& 1  & 0.73& 175\\
        & 2         & 0.95& 244\\
        & \textbf{3 }       & \textbf{0.96}& 256\\
        
        \midrule
        \multirow{3}{*}{GLM-5.1} & 1  & 0.71& 219\\
        & 2         &0.88& 305\\
        & \textbf{3}       & \textbf{0.91}& 348\\

        \midrule
        \multirow{3}{*}{DeepSeek-v4}& 1  & 0.44& 162\\
        & 2         & 0.64& 288\\
        & \textbf{3}       & \textbf{0.75}& 385\\
        
        \bottomrule
    \end{tabular}
    
     \caption{Effect of iterative refinement epochs on workflow synthesis performance. Increasing epochs enables additional closed-loop repair but incurs higher token consumption.}
     \label{tab:model_ab_epoch}
\end{table}

\subsection{Main Results}
\label{ssec:main_results}
Table~\ref{tab:model_evaluation} shows that direct prompting is insufficient for production-grade Dify synthesis. ZeroShot reaches 0.43 SR with Qwen-3.7 and only 0.13--0.15 with GLM-5.1 and DeepSeek-v4; adding dynamic repair improves SR to 0.67, 0.45, and 0.36, but remains far below the full system.
\textsc{ChatDify} achieves 0.96, 0.91, and 0.75 SR across the three backends, improving over ZeroShot+Dynamic by 29, 46, and 39 percentage points respectively. This shows that reliability comes from the full Planner--Executor--Reviewer pipeline rather than runtime repair alone. Qwen-3.7 gives the strongest accuracy--cost profile (0.96 SR with 256k tokens), while DeepSeek-v4 is less effective on this benchmark. Across all backends, input tokens account for 80--90\% of the reported totals, as retrieved knowledge base entries and the workflow DSL carried across refinement epochs accumulate in the prompt.

Table~\ref{tab:model_ab_sd} shows that knowledge grounding alone is insufficient: it guides component selection but cannot verify whether a composed workflow executes correctly or fulfills the request. Removing dynamic probing therefore sharply reduces SR (0.96 to 0.73, 0.91 to 0.71, and 0.75 to 0.44), while static verification adds a complementary 3--8 percentage point gain by filtering structural errors before execution. The lowest SR without both stages (0.60 and 0.33 for GLM-5.1 and DeepSeek-v4) validates their complementary roles in reliable synthesis.

Finally, Table~\ref{tab:model_ab_epoch} shows the cost--quality trade-off of refinement. Most gains appear within two epochs (Qwen-3.7: 0.73 to 0.95; GLM-5.1: 0.71 to 0.88; DeepSeek-v4: 0.44 to 0.64), while the third epoch brings smaller gains at higher token cost. Thus, two epochs suit high-throughput deployment, whereas three epochs favor maximum success rate.

\section{Conclusions}

We introduced \textsc{ChatDify}, a multi-agent framework for synthesizing Dify workflows from natural language. Its Planner--Executor--Reviewer architecture bridges user intent and platform-specific DSL configuration through knowledge-grounded planning, schema-aware compilation, and hybrid verification. On 100 production-inspired tasks, \textsc{ChatDify} consistently outperforms practical zero-shot baselines, validating its value for industrial Dify workflow automation. Although our implementation targets Dify, the core design is extensible to other workflow orchestration platforms that expose structured node schemas, executable DSLs, and runtime validation interfaces.

\section*{Limitations}

\textsc{ChatDify} is implemented and evaluated on Dify, which allows us to study NL-to-workflow synthesis under a concrete industrial DSL with real deployment constraints. This focus also means that direct transfer to other workflow platforms requires adapting the knowledge base, node schemas, and sandbox executor to the target platform. However, these components are modular in our Planner--Executor--Reviewer design, and the same principles apply to platforms that expose structured node specifications and executable workflow definitions.

Our benchmark contains 100 production-inspired tasks, covering major Dify workflow categories and common application patterns. While this scale is sufficient to compare design choices and baselines, broader cross-organization benchmarks would further improve external validity. In addition, dynamic probing and LLM-as-Judge verification introduce extra token cost, reflecting an engineering trade-off between reliability and inference budget. Our epoch analysis suggests that this cost can be tuned in practice: two refinement rounds provide strong performance for high-throughput settings, while additional rounds can be reserved for higher-stakes workflows.

Finally, SR measures execution success together with LLM-judged semantic success, and does not fully reflect human judgment of task completion. We adopt this automated protocol because open-ended workflow outputs, such as generated reports or replies, lack deterministic ground truth, and our experiments produce thousands of workflows across backbones, runs, baselines, and ablations, which makes exhaustive human evaluation impractical. Reusing the Reviewer's verifier as the judge also keeps the evaluation aligned with the checks that drive refinement. The cost of this choice is that each check relies on a single synthesized test case and the judge shares its backbone LLM with the generator, so SR may overestimate the true success rate. Although our manual spot checks found that most workflows behaved as expected, a systematic human evaluation with multiple test cases and independent annotators would provide a stricter assessment.

\bibliography{ref}

\appendix

\section{Dify Workflow Example}

\begin{figure}[h]
    \centering
    \subfigure[Workflow interface demonstrating complex research task automation]{
      \includegraphics[width=1\linewidth]{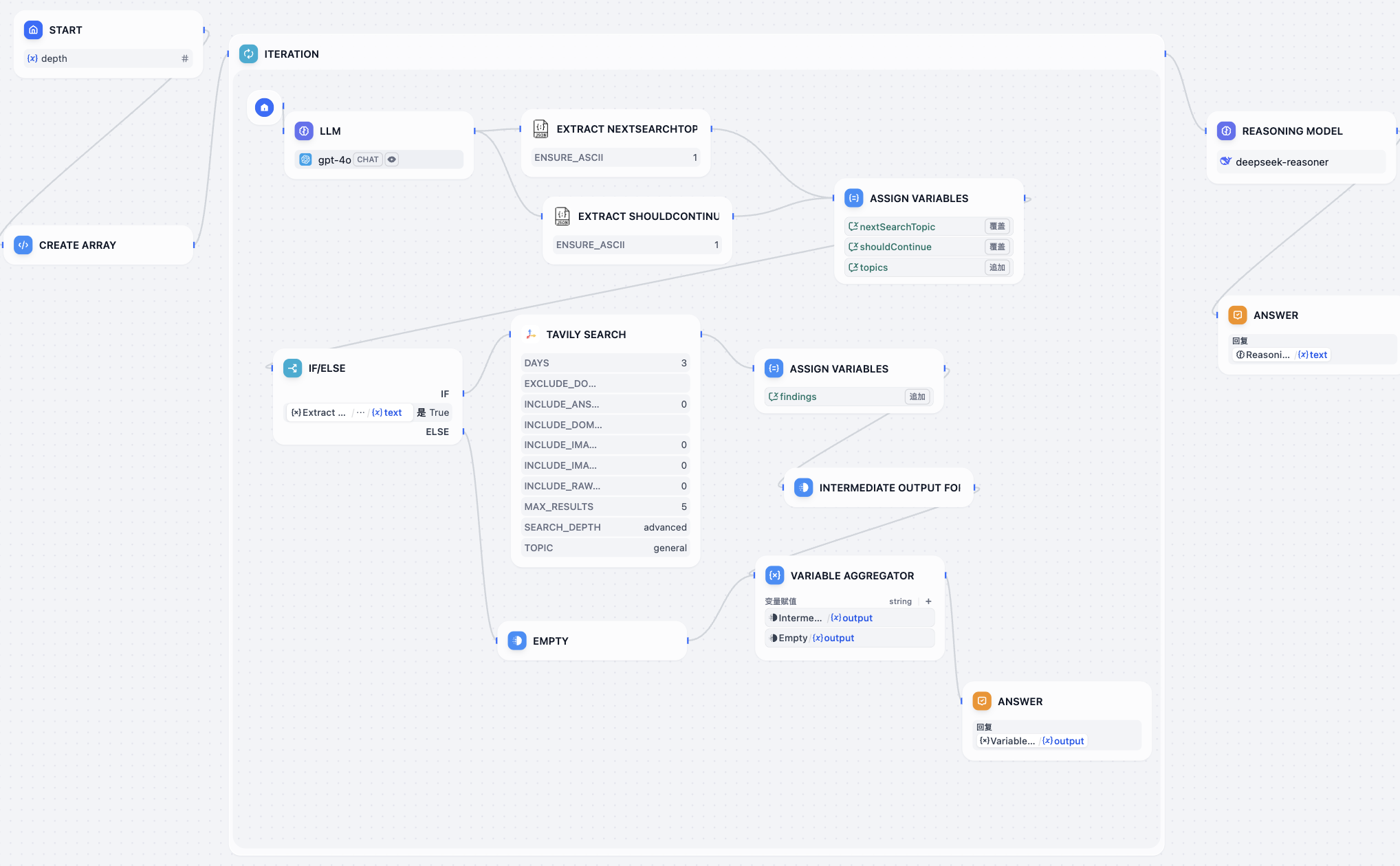}
      \label{fig:dify_deepresearch_demo}
    }
    \subfigure[Partial YAML configuration]{
      \includegraphics[width=1\linewidth]{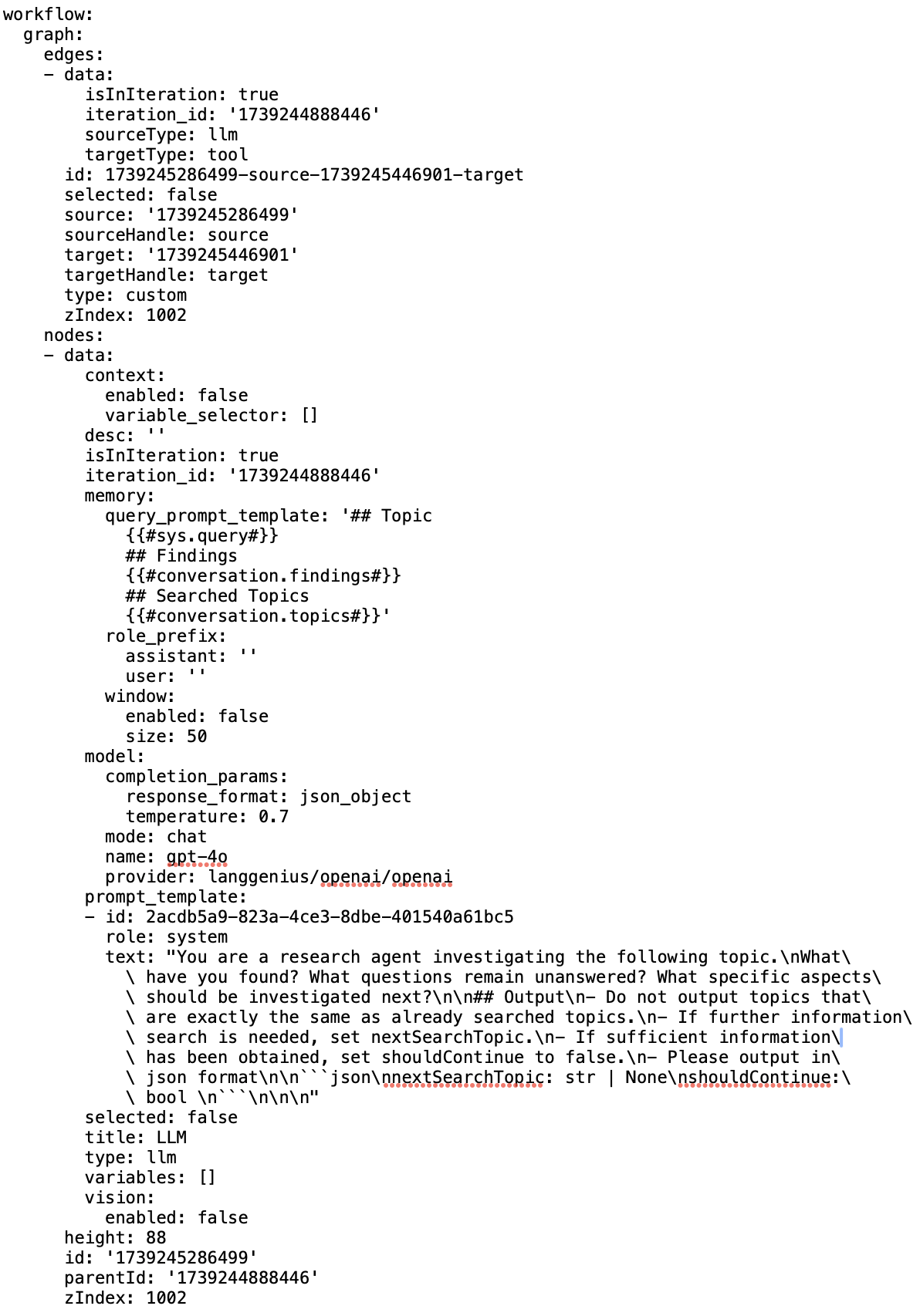}
      \label{fig:dify_deepresearch_yaml_demo}
    }
    \caption{Example of Dify DeepResearch workflow (from dify.ai): (a) workflow interface and (b) partial YAML configuration.}
    \label{fig:dify_deepresearch}
  \end{figure}

This section presents an example of complex Dify workflows to illustrate the structure and complexity of real-world workflow automation on the Dify platform. We use Dify DeepResearch as a representative example, which demonstrates how sophisticated research tasks can be orchestrated through a visual workflow interface. The Figure~\ref{fig:dify_deepresearch} shows the workflow interface and its corresponding YAML configuration.

\section{Dify Knowledge Base}
\label{app:node_kb}

The Dify Knowledge Base serves as a comprehensive repository that provides structured specifications and exemplars for all available Dify nodes and their interconnections.  The knowledge base is organized into four complementary components, each serving a distinct purpose in the workflow synthesis process.

\subsection{Edge Usage Specifications}

\begin{figure}[h]
  \centering
 \includegraphics[width=1.0\linewidth]{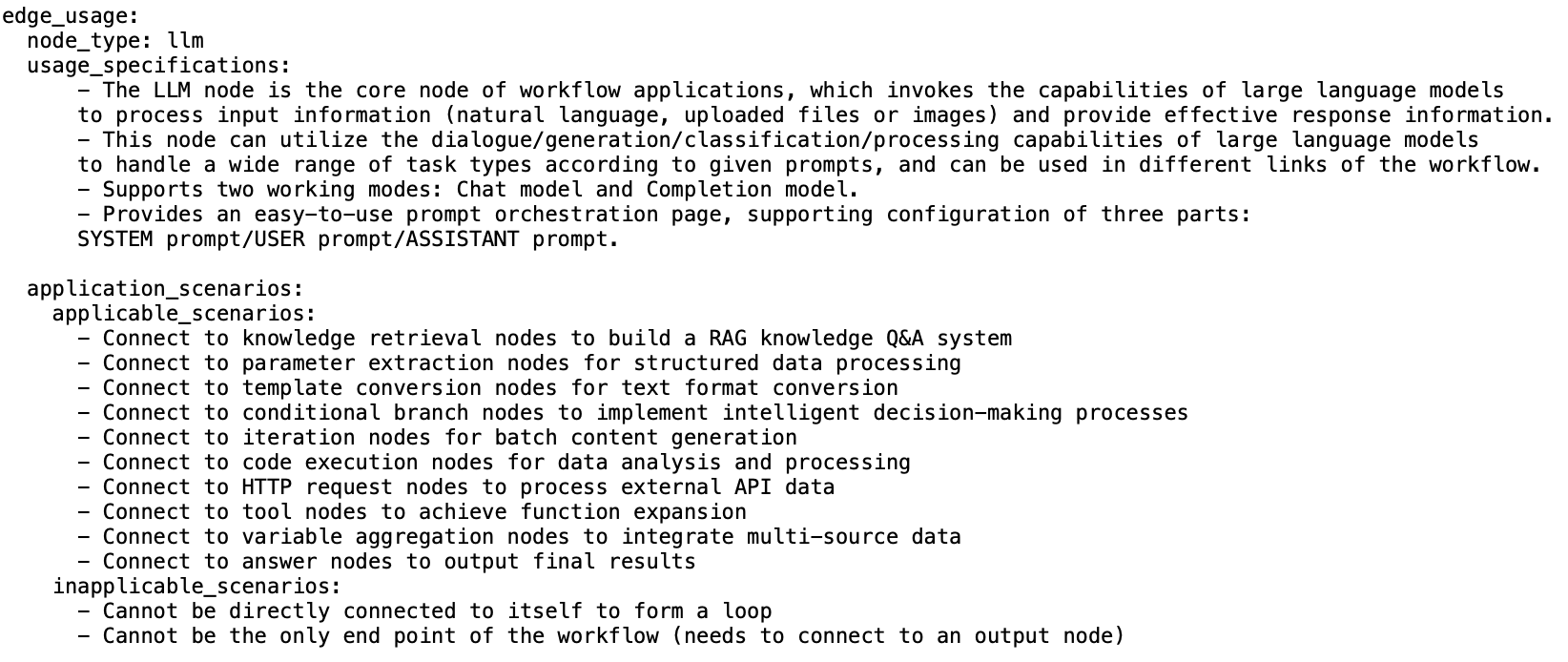} 
\caption{Example of Edge Usage Specifications for LLM node type}
\label{fig:edge_usage_example}
\end{figure}

The \textbf{Edge Usage Specifications} component defines how nodes can be connected within workflows. For each node type, it provides:

\begin{itemize}
    \item \textbf{Usage Specifications}: A description of the node's role and capabilities within workflow applications, including its primary functions and supported operation modes.
    
    \item \textbf{Application Scenarios}: A structured categorization of connection patterns:
    \begin{itemize}
        \item \textbf{Applicable Scenarios}: Valid connection patterns that represent common and recommended usage patterns. For example, LLM nodes can connect to knowledge retrieval nodes for RAG systems, parameter extraction nodes for structured data processing, conditional branch nodes for decision-making, and various other node types depending on the task requirements.
        
        \item \textbf{Inapplicable Scenarios}: Invalid or problematic connection patterns that should be avoided. These include direct self-loops, incomplete workflow structures (e.g., LLM nodes that are the only endpoint without connecting to output nodes), and other patterns that violate Dify's workflow constraints.
    \end{itemize}
\end{itemize}
This component guides the \textit{Planner Agent} in constructing valid execution plans and helps the \textit{Reviewer Agent} detect structural issues during static verification.
Figure~\ref{fig:edge_usage_example} presents an example of edge usage specifications for LLM nodes, detailing their connection patterns and constraints.

\subsection{Edge Examples}

The \textbf{Edge Examples} component contains validated edge configurations that demonstrate correct node-to-node connections. These examples illustrate proper data flow patterns and help ensure that the generated workflows maintain correct topological ordering and variable scoping.

\begin{figure}[h]
  \centering
 \includegraphics[width=0.6\linewidth]{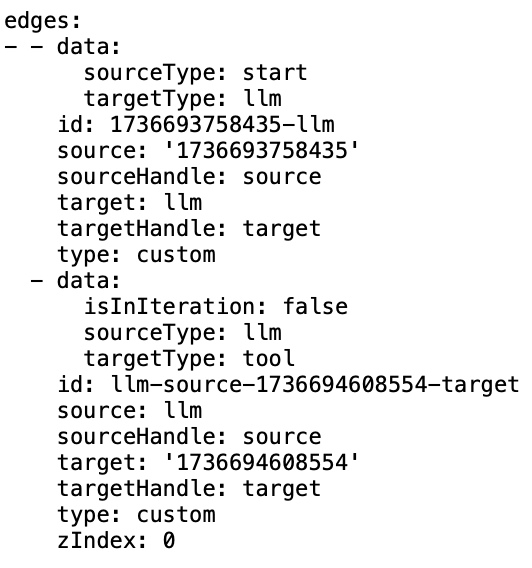} 
\caption{Example of Edge Configurations showing valid node connections}
\label{fig:edge_example}
\end{figure}

As an illustration, Figure~\ref{fig:edge_example} demonstrates valid edge configurations, showing connections from a start node to an LLM node and from an LLM node to a tool node, including source and target node types, edge metadata, connection handles, and contextual information.

\subsection{Node Usage Specifications}

\begin{figure}[h]
  \centering
 \includegraphics[width=1.0\linewidth]{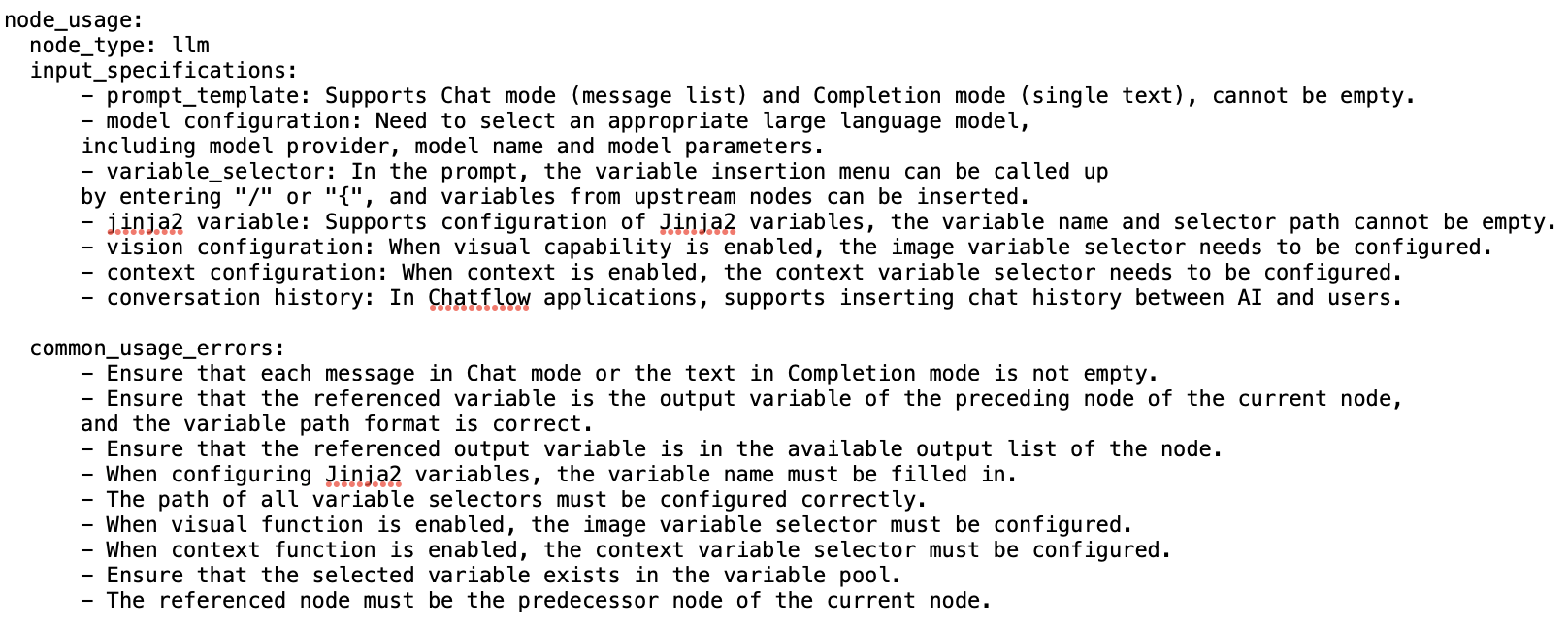} 
  \caption{Example of Node Usage Specifications for LLM node type}
  \label{fig:node_usage_example}
\end{figure}

The \textbf{Node Usage Specifications} component provides detailed guidelines for configuring each node type. For each node type, it includes:

\begin{itemize}
    \item \textbf{Input Specifications}: A comprehensive list of required and optional input parameters, their data types, constraints, and configuration requirements. For instance, for LLM nodes, this includes specifications for prompt templates (supporting both Chat and Completion modes), model configuration (provider, model name, and parameters), variable selectors, Jinja2 variables, vision and context configurations, and conversation history settings.
    
    \item \textbf{Common Usage Errors}: A curated list of frequently encountered configuration mistakes and validation rules. These error patterns help the \textit{Reviewer Agent} identify potential issues during static verification. Examples include ensuring non-empty prompt templates, validating variable references against the variable pool, verifying correct variable path formats, and checking that referenced nodes are valid predecessors in the workflow graph.
\end{itemize}
This component enables the \textit{Executor Agent} to make informed decisions about parameter initialization and helps prevent common configuration errors during workflow synthesis.
Figure~\ref{fig:node_usage_example} illustrates a concrete example of node usage specifications for an LLM node type.

\subsection{Node Examples}

The \textbf{Node Examples} component contains concrete, validated node configurations in Dify's DSL format. Each example demonstrates a complete node instantiation with all required fields populated according to the node type. These examples serve as templates that the \textit{Executor Agent} can adapt based on the user's requirements and the structured plan from the \textit{Planner Agent}. By retrieving semantically similar examples from this component, the Executor can initialize node parameters more accurately, reducing the need for iterative refinement.

\begin{figure}[h]
  \centering
 \includegraphics[width=1.0\linewidth]{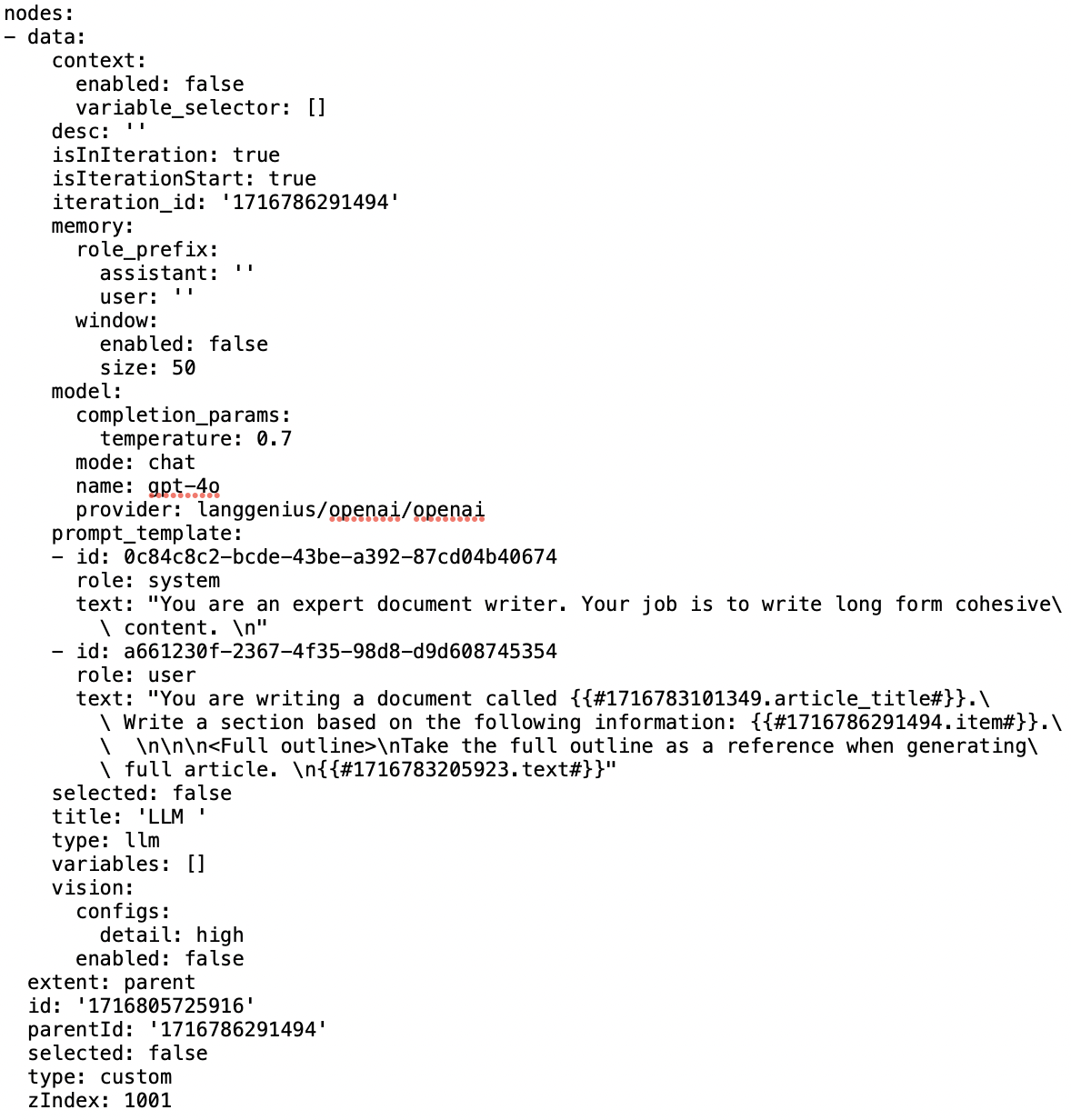} 
\caption{Example of Node Configuration for an LLM node within an iteration block}
\label{fig:node_example}
\end{figure}

As an illustration, Figure~\ref{fig:node_example} shows a complete node configuration example for an LLM node used within an iteration block, including model configuration, prompt templates, variable bindings, and structural properties.

\subsection{Knowledge Base Integration}

The four components work together to support the multi-agent workflow synthesis process. During planning, the \textit{Planner Agent} queries the Edge Usage Specifications and Edge Samples to construct valid execution plans. During execution, the \textit{Executor Agent} retrieves relevant Node Examples and adapts them based on Node Usage Specifications to instantiate configured nodes. The \textit{Reviewer Agent} leverages all four components to perform comprehensive validation: it uses Node Usage Specifications to check parameter correctness, Edge Usage Specifications to verify structural validity, and compares the generated workflow against the examples to identify deviations from established patterns.

This structured knowledge representation enables \textsc{ChatDify} to generate workflows that are not only syntactically correct but also follow Dify's best practices and common usage patterns, significantly improving the success rate and quality of generated workflows.

\section{Case Study}

We present several representative case studies that demonstrate \textsc{ChatDify}'s capability to handle diverse workflow requirements.

\subsection{Competitive Analysis Report}

\textbf{User Request:} "Generate an analysis and a competitive report for three mainstream AI Agent tools, namely Dify, Coze and N8n. The report shall cover the following comparison dimensions: core positioning, key technical features, learning curve, deployment and data security, pricing model, strengths and weaknesses, typical applicable scenarios, etc. The output format of the research report shall be a Markdown document."

\begin{figure*}[h]
  \centering
 \includegraphics[width=1.0\linewidth]{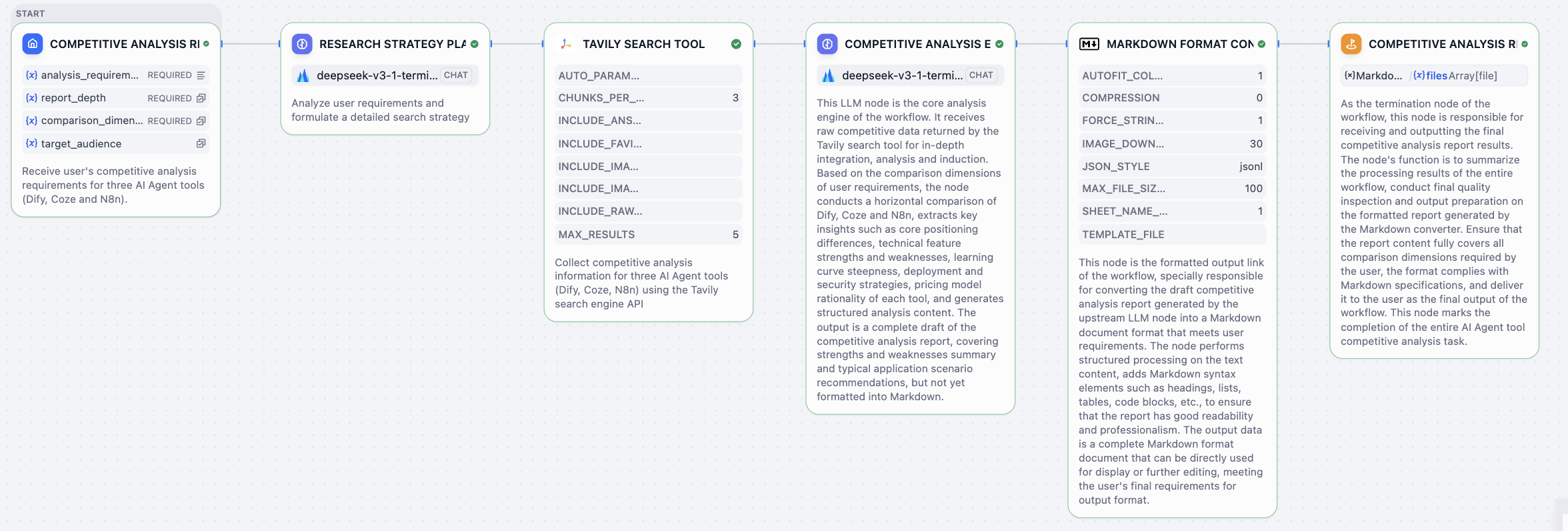} 
  \caption{\textsc{ChatDify}-synthesized dify workflow for competitive analysis report generation.}
  \label{fig:competitive_analysis}
\end{figure*}

\textbf{Synthesized Dify Workflow:} \textsc{ChatDify} synthesizes a sequential workflow: (i) a start node that receives user inputs including analysis requirements, report depth, comparison dimensions, and target audience, (ii) an LLM node for research strategy planning, (iii) a Tavily search tool node for information collection, (iv) an LLM node for competitive analysis, (v) a Markdown converter tool node for report formatting, and (vi) an end node that outputs the final competitive analysis report. The workflow is shown in Figure~\ref{fig:competitive_analysis}.

\subsection{Public Opinion Report with Iteration}

\textbf{User Request:} "Generate a public opinion report based on the latest news relevant to the user’s keywords. For each news article, extract its title and conduct sentiment analysis, then compile the findings into a comprehensive report."

\begin{figure*}[h]
  \centering
 \includegraphics[width=1.0\linewidth]{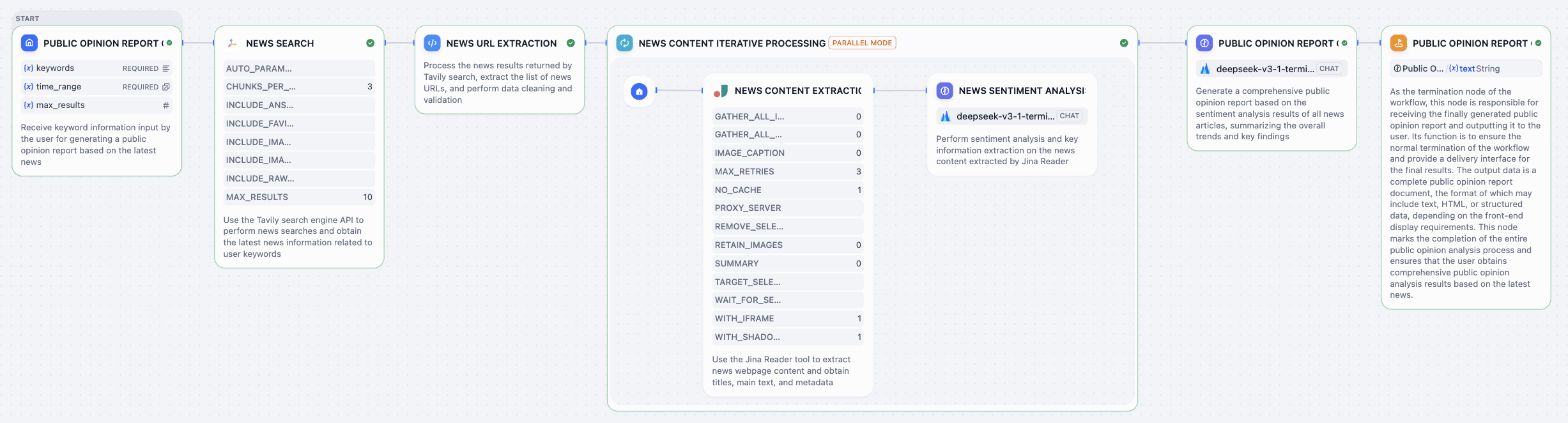} 
  \caption{\textsc{ChatDify}-synthesized dify workflow for public opinion report generation.}
  \label{fig:public_opinion}
\end{figure*}

\textbf{Synthesized Dify Workflow:} \textsc{ChatDify} synthesizes a workflow with iterative processing: (i) a start node that receives user inputs including keywords, time range, and maximum number of news articles, (ii) a Tavily search tool node for retrieving relevant news articles, (iii) a code node that extracts and validates news URLs from the search results, (iv) an iteration node that processes each news URL in parallel, containing a Jina Reader tool node for content extraction and an LLM node for sentiment analysis, (v) an LLM node that aggregates all sentiment analysis results to generate a comprehensive public opinion report, and (vi) an end node that outputs the final report. The workflow is shown in Figure~\ref{fig:public_opinion}.

\subsection{Intelligent Contract Analysis}

\textbf{User Request:} "Generate an intelligent contract analysis workflow that supports batch processing of contract documents, automatically identifies risks, and generates analysis reports. The workflow should iteratively process all uploaded contract documents. For each document, it should: (i) identify and verify whether it is a contract-type file (non-contract files should be directly prompted and the process terminated), (ii) automatically classify contract types (procurement, labor, lease, service, etc.), (iii) extract key information including contract subject information (contracting parties, contact information, qualifications), core clauses (rights and obligations, payment methods, delivery standards, acceptance conditions), critical dates (signing date, effective date, expiration date, payment deadlines, delivery deadlines), and financial information (total contract amount, payment ratios, penalty amounts), (iv) perform risk detection and rating including clause completeness checks, potential risk identification, compliance checks, critical date reminders, and risk level assessment (high/medium/low risk). Finally, generate an overall analysis report including contract quantity, contract type distribution, risk level distribution, and risk point distribution in Markdown document format."

\begin{figure*}[h]
  \centering
 \includegraphics[width=1.0\linewidth]{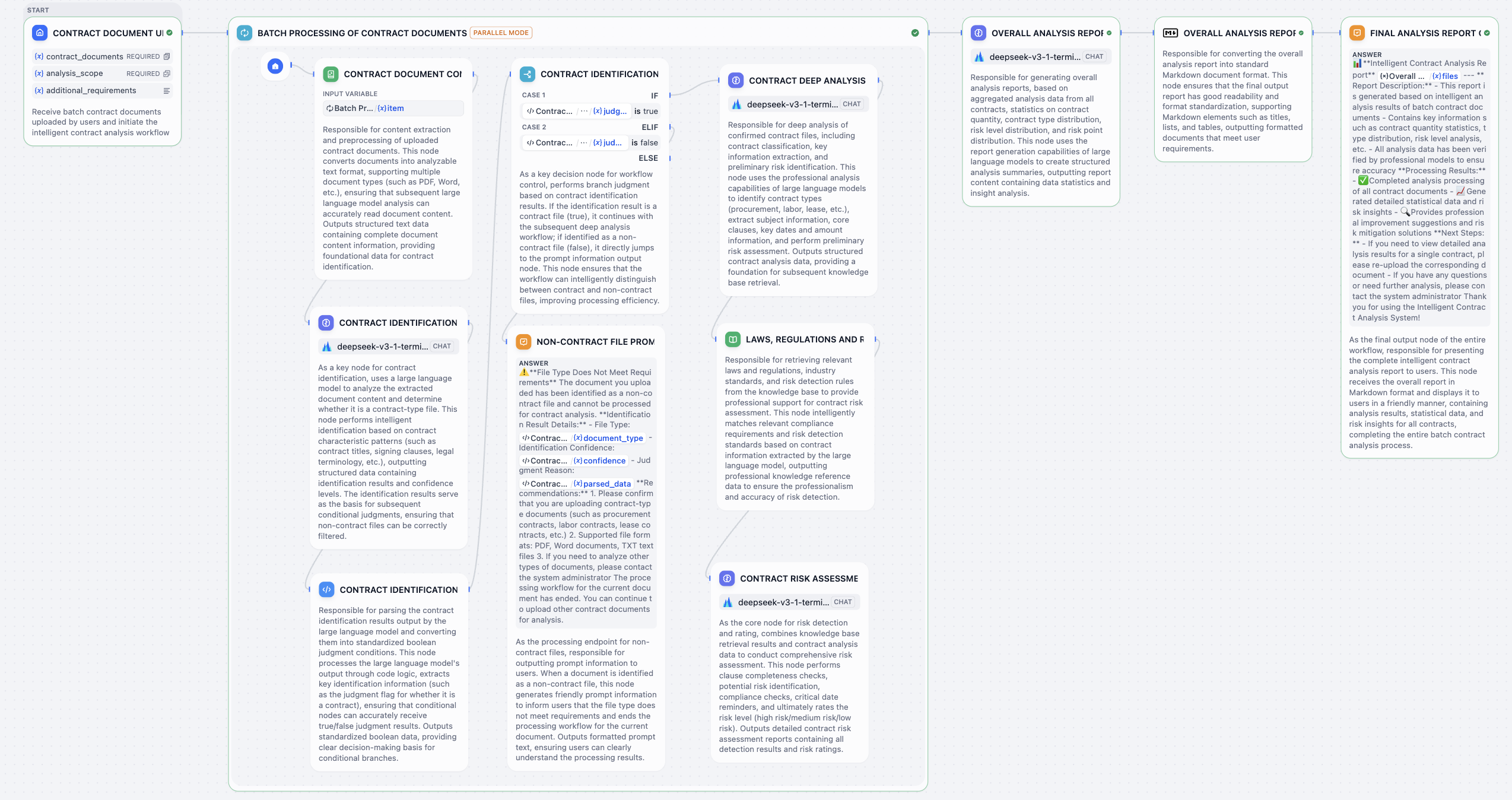} 
  \caption{\textsc{ChatDify}-synthesized dify workflow for intelligent contract analysis.}
  \label{fig:contract_analysis}
\end{figure*}

\textbf{Synthesized Dify Workflow:} \textsc{ChatDify} synthesizes a workflow with iterative batch processing: (i) a start node that receives user inputs including contract documents, analysis scope, and additional analysis requirements, (ii) an iteration node that processes each contract document in parallel, containing: (a) a document extractor node for content extraction, (b) an LLM node for contract identification analysis, (c) a code node that parses the identification results and converts them to boolean judgment conditions, (d) an if-else node that branches based on contract identification results, (e) for contract files: an LLM node for deep contract analysis (classification, key information extraction, and preliminary risk identification), a knowledge retrieval node for retrieving relevant laws, regulations, and risk detection rules, an LLM node for comprehensive risk assessment and rating, and a variable aggregator node for collecting analysis results, (f) for non-contract files: an answer node that outputs a prompt message, (iii) an LLM node that generates an overall analysis report based on aggregated contract analysis data, (iv) a Markdown formatting tool node that converts the report to standard Markdown format, and (v) an end node that outputs the final intelligent contract analysis report. The workflow is shown in Figure~\ref{fig:contract_analysis}.

\end{document}